\documentclass[lettersize,journal]{IEEEtran}
\usepackage{amsmath,amsfonts}
\usepackage{algorithmic}
\usepackage{algorithm}
\usepackage{array}
\usepackage[caption=false,font=normalsize,labelfont=sf,textfont=sf]{subfig}
\usepackage{textcomp}
\usepackage{stfloats}
\usepackage{url}
\usepackage{verbatim}
\usepackage{graphicx}
\usepackage{cite}
\usepackage{booktabs}
\usepackage{multirow}
\usepackage{xcolor}

\begin{document}

\title{Robust and Generalizable GNN Fine-Tuning via Uncertainty-aware Adapter Learning}

 \author{Bo Jiang, Weijun Zhao, Beibei Wang, Xiao Wang, Jin Tang 
 \thanks{
The authors are with School of Computer Science and Technology, Anhui University. 
 Manuscript received April 19, 2021; revised August 16, 2021.}
}

\markboth{Journal of \LaTeX\ Class Files,~Vol.~14, No.~8, August~2021}%
{Shell \MakeLowercase{\textit{et al.}}: A Sample Article Using IEEEtran.cls for IEEE Journals}


\maketitle

\begin{abstract}
Recently, fine-tuning large-scale pre-trained GNNs has yielded remarkable attention in adapting pre-trained GNN models for downstream graph learning tasks. 
One representative fine-tuning method is to exploit adapter (termed AdapterGNN) which aims to `augment' the pre-trained model by inserting 
a lightweight module to make the `augmented' model better adapt to the downstream tasks. 
However, graph data may contain various types of noise in downstream tasks, such as noisy edges and ambiguous node attributes. 
Existing AdapterGNNs are often prone to graph noise and exhibit limited generalizability.  
\emph{How to enhance the robustness and generalization ability of GNNs' fine-tuning remains an open problem. }

In this paper, we show that the above problem can be well addressed by integrating uncertainty learning into the GNN adapter.
We propose the Uncertainty-aware Adapter (UAdapterGNN) that fortifies pre-trained GNN models against noisy graph data in the fine-tuning process. 
Specifically, in contrast to regular AdapterGNN, our UAdapterGNN
exploits Gaussian probabilistic adapter to augment the pre-trained GNN model. In this way, when the graph contains various noises, our method can automatically absorb the effects of changes in the variances of the Gaussian distribution, thereby significantly enhancing the model's robustness. Also, UAdapterGNN can further improve the generalization ability of the model on the downstream tasks. Extensive experiments on several benchmarks demonstrate the effectiveness, robustness and high generalization ability of the proposed UAdapterGNN method.  
\end{abstract}


\section{Introduction}
Recently, fine-tuning paradigms have emerged as pivotal 
techniques to achieve knowledge transfer from pre-trained GNN models to the downstream graph learning tasks~\cite{wu2020comprehensive}. 
Current research in this field can be broadly categorized into three types, i.e., full fine-tuning, prompt-based fine-tuning, and adapter-based fine-tuning. 
Full fine-tuning~\cite{xie2022self, hu2019strategies} aims to first leverage transferable representations learned from unlabeled graphs
and 
subsequently adapt them  to the downstream tasks by updating (often all) model parameters. 
However, this paradigm
is usually computationally expensive in terms of memory and time. 
The prompt-based fine-tuning approach freezes the pre-trained GNN model and designs some learnable graph prompts on each downstream task, which are directly appended to the input to enable adaptation. 
For example, GPF~\cite{gpf} adds a global learnable prompt to downstream graph data, achieving performance comparable to full fine-tuning. 
Similarly, EdgePrompt~\cite{Fu2025EdgePT}  introduces learnable prompt vectors on edges to better aggregate with node representations during message passing.
GPPT~\cite{sun2022gppt} incorporates a `label-node' pair prompt into the input, unifying downstream node classification with the pre-trained link prediction task. 
GraphPrompt~\cite{liu2023graphprompt} further unifies the pre-training and downstream stages by using a single task template. 
In addition to prompt learning, adapter-based fine-tuning offers an alternative PEFT strategy, which aims to insert lightweight trainable modules (\textit{adapters}) into the frozen backbone of the pre-trained model, enabling adaptation with minimal parameter updates. 
Recent advancements in graph adapters have introduced diverse strategies to align pre-trained models with downstream tasks. 
For example, G-Adapter~\cite{g-adapter} encodes topology via Bregman divergence to mitigate feature distribution shifts caused by domain discrepancies.  
GConv-Adapter~\cite{papageorgiougraph} combines dual-normalized convolution and low-rank matrices to capture higher-order information efficiently.
GraphLoRA~\cite{yang2024graphlora} injects low-rank adaptation modules, enabling efficient cross-graph domain transfer. 
AdapterGNN~\cite{adaptergnn} proposes to adapt to downstream tasks effectively by leveraging expressive adapter modules with minimal trainable parameter, simultaneously significantly enhancing the model's generalization capabilities. 

However, in many downstream learning tasks, graphs may contain various kinds of noises, such as noisy edges and ambiguous node attributes. 
After reviewing the above adapters, we observe that existing works mainly focus on efficient knowledge transfer from pre-trained GNN models to the downstream tasks. 
However, \textbf{little attention has been devoted  to robust learning on downstream tasks with respect to graph noises}. 
In this paper, we show that existing adapter approaches generally adopt deterministic adapter modules, which are prone to being sensitive to graph noise and also have limited generalization ability. 
\textbf{Therefore, how to enhance the robustness and generalization ability of GNNs' fine-tuning remains an open problem in recent years.}


In this paper, we show that the above problem can be well addressed by integrating uncertainty learning into GNN adapter.  
Specifically, we propose the Uncertainty-aware Adapter (UAdapterGNN), a novel framework that enhances large pre-trained GNNs against noisy graph data in the downstream tasks. 
%
Instead of designing a deterministic adapter, our UAdapterGNN method exploits Gaussian random adapter to augment the pre-trained GNN model. 
%
In this way, when the graphs contain noises, the proposed UAdapterGNN can automatically absorb the effects of changes in the variances of the Gaussian distributions, thereby significantly enhancing the pre-trained model’s robustness on the downstream tasks. 
Moreover, it can improve the model’s generalization ability on downstream tasks when compared to traditional deterministic adapter models. 

Uncertainty learning methods have been usually studied in machine learning field~\cite{gal2016dropout,kendall2017uncertainties}. We note that it has also been exploited in fine-tuning pre-trained LLM/VLM tasks in recent years~\cite{WangZ0ZLT24,JiWGZZWZSY23}.  
However, to our best knowledge, this work is the first attempt to exploit uncertainty learning model into GNN's fine-tuning problem. 
More importantly, we show that the uncertainty adapter learning can well improve the robustness and generalization of pre-trained GNN model w.r.t noisy data on downstream tasks. 
Note that, the proposed UAdapterGNN method offers a general \emph{plug-and-play adapter} architecture
 which can be integrated with any pre-trained GNN models to enable \textbf{robust learning} on downstream tasks. 
 Overall, the main contributions of this paper are summarized as follows,
\begin{itemize}
\item We propose to leverage 
uncertainty learning model into 
GNN fine-tuning problem to enhance the robustness and generalization ability of pre-trained GNN models w.r.t noisy data in the downstream tasks. 
\item We design a lightweight 
architecture to implement our 
 UAdapterGNN by extending 
 traditional AdapterGNN model into uncertainty learning manner, as demonstrated in Figure~\ref{fig:framework}.
 
 \item 
We evaluate UAdapterGNN on several benchmark datasets. Experimental results show the effectiveness and advantages (robustness, generalization ability) of our method under different GNN pre-training strategies. 
\end{itemize}

\section{Related Work}

\textbf{Pre-training on Graphs.} 
To mitigate reliance on labeled data in supervised learning, pre-training GNNs has become increasingly popular, inspired by pre-training techniques in language and vision applications~\cite{HuLZHNL24,cv}.
These methods train graph encoders via self-supervised tasks~\cite{LiuJPZZXY23} to learn transferable representations, thereby improving cross-task generalization capabilities.
For example, in work~\cite{hu2019strategies}, it proposes a pre-training strategy for Graph Neural Networks (GNNs), which effectively enhances the model's generalization ability and accuracy across various downstream tasks by combining node-level and graph-level pre-training approaches. Graph Contrastive Learning~\cite{you2020graph} employs graph augmentation-based contrastive learning to pre-train GNNs, facilitating the extraction of invariant representations against structural perturbations. GPT-GNN~\cite{GPT-GNN} proposes to pre-train GNNs through self-supervised generative tasks, enabling them to jointly capture structural patterns and node semantic attributes in graphs. 
GCC~\cite{GCC} introduces contrastive graph context prediction as its pre-training task, empowering GNNs to capture transferable structural motifs across diverse network domains.

\textbf{Graph Adapter-Tuning.} 
Pre-trained graph models leverage self-supervised learning to capture universal graph structural knowledge, yet their full-parameter fine-tuning suffers from high computational overhead and overfitting risks on downstream tasks. To address this, the `pre-train, adapter-tuning' paradigm has emerged, which inserts adapter modules with lightweight neural network architecture to bridge the knowledge transfer between pre-trained models and downstream tasks~\cite{adapter}.  
For example, G-Adapter~\cite{g-adapter} addresses the critical challenge of feature distribution shift in graph neural network transfer learning by integrating graph topological relation encoding and Bregman divergence-based optimization strategies. GConv-Adapter~\cite{papageorgiougraph} combines dual-normalized graph convolution and trainable low-rank weight matrices. It effectively captures second-order adjacency information which enhances structural awareness and enables parameter-efficient fine-tuning. 
VRGAdapter~\cite{jiang2025beyond} proposes a parameter-efficient adapter that generates multiple textual descriptions per class, encodes them as Gaussian-distributed nodes in a random knowledge graph, and uses reparameterization to sample the adapter representations. AdapterGNN~\cite{adaptergnn} proposes a parameter-efficient fine-tuning adapter architecture to address the problem of insufficient generalization capability and excessive computational resource consumption existed in traditional full-parameter fine-tuning methods. 

\section{Preliminaries}

Let \(f_\Omega:\mathcal{G} \mapsto \mathbb{R}^d\) denote a pre-trained GNN model with frozen parameters \(\Omega\) and \(\mathcal{G}=(\mathcal{V, E}, \mathbf{X},\mathbf{A})\) denote an input downstream graph with \(\mathcal{V}\) and \(\mathcal{E}\) representing the set of nodes and edges, respectively. Here, \(\mathbf{X}=(\mathbf{x}_1, \mathbf{x}_2\cdots \mathbf{x}_N)\in \mathbb{R}^{N \times d}\) represents the collection of node features and \(\mathbf{A} \in \{0,1\}^{N \times N}\) represents the adjacency matrix.  
The message passing in the $l$-th layer of pre-trained GNN model can be usually formulated as follows,
\begin{equation}
{\mathbf{y}}_i^{(l)} =\mathrm{MLP} \Big( \bigoplus_{j \in \mathcal{N}(i)} m^{(l)} \big( {\mathbf{x}}_i^{(l)}, {\mathbf{x}}_j^{(l)}, {e}_{ji} \big) \Big)
\label{eq: massage_passing}
\end{equation}
where 
$\mathcal{N}(i)$ indicates the neighborhood of node $v_i$ and $\bigoplus$ denotes a permutation-invariant aggregation operator (typically summation or mean). $m^{(l)}$ represents the message function. All parameters $\Omega$ in above message passing operation are frozen during fine-tuning. 

\begin{figure*}[!htbp]
  \centering
  \subfloat[\footnotesize Regular Adapter in GNN]{
    \includegraphics[width=0.42\linewidth]{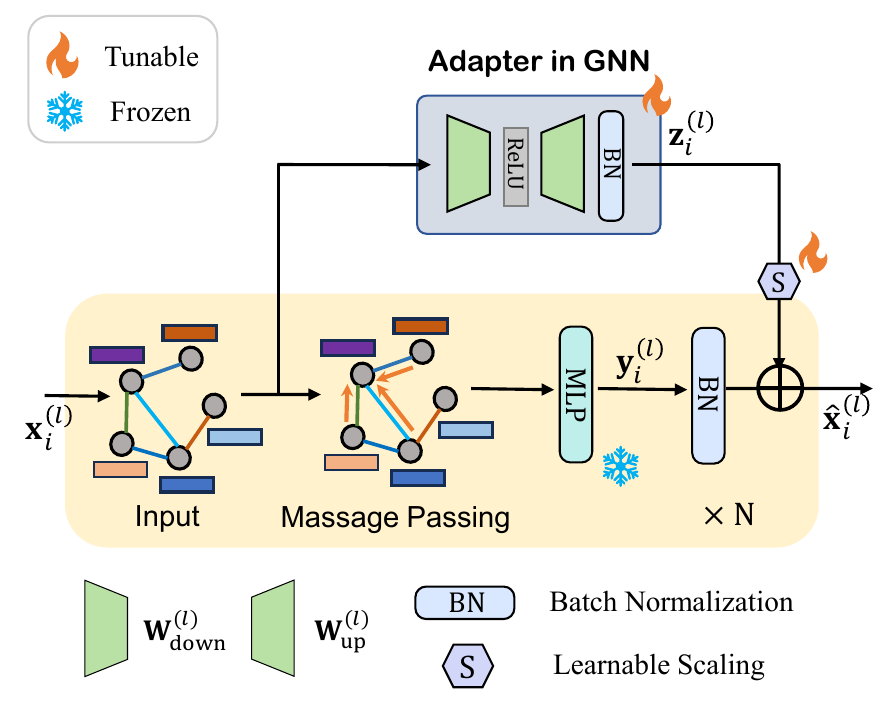}
    
  }
  \hspace{1cm}
  \subfloat[\footnotesize Our proposed UAdapterGNN]{
    \includegraphics[width=0.42\linewidth]{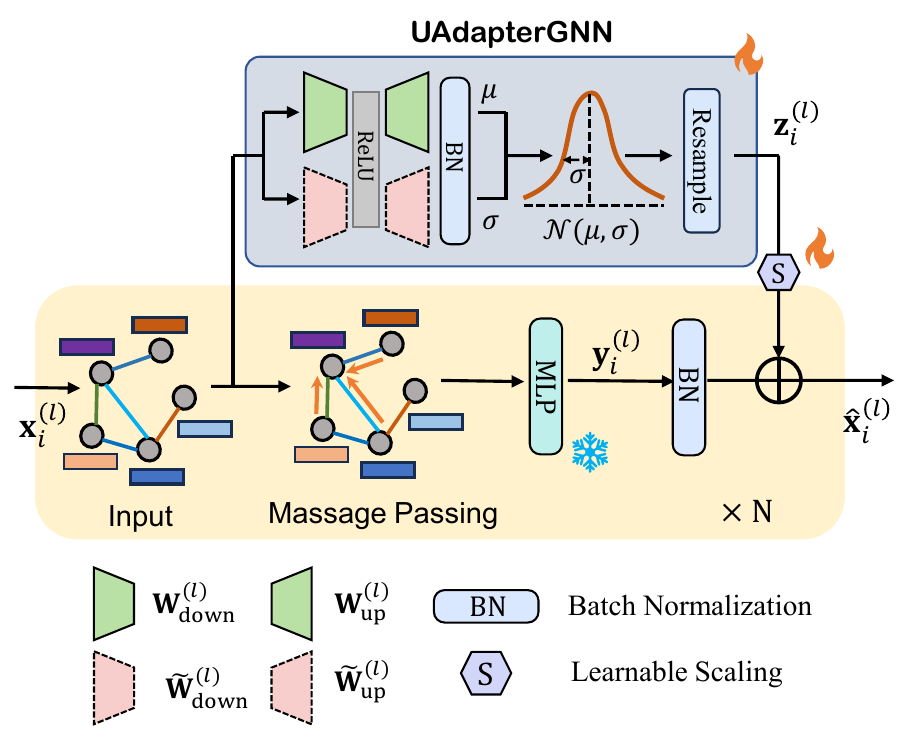}
  }
  \caption{Comparison of regular Adapter in GNN and our UAdapterGNN. In each  layer of the pre-trained GNN model, UAdapterGNN learns a Gaussian module to argument the pre-trained GNN backbone on downstream tasks. 
  }
  \label{fig:framework}
\end{figure*}

The Adapter module in GNN projects 
the $l$-th layer input \(\mathbf{x}^{(l)}_i \in \mathbb{R}^{d_{\mathrm{in}}}\)
into a low-dimensional space via a down-projection matrix \(\mathbf{W}_{\mathrm{down}}\in\mathbb{R}^{d_{\text{mid}}\times d_{\text{in}}}\), followed by a $\mathrm{ReLU}$ activation and up-projection matrix \(\mathbf{W}_{\mathrm{up}}\in\mathbb{R}^{d_{\text{out}}\times d_{\text{mid}}}\), where \(d_{\text{mid}} \ll d_{\text{in}}\). 
To stabilize training and mitigate distribution shifts, it also incorporates trainable Batch Normalization (BN) after the up-projection. Formally, adapter outputs $\mathbf{z}^{(l)}_i$ as, 
\begin{equation}
    \mathbf{z}^{(l)}_i =\mathrm{BN}\left(\mathbf{W}^{(l)}_{\text{up}} \cdot \mathrm{ReLU}\left(\mathbf{W}^{(l)}_{\text{down}} \cdot \mathbf{x}^{(l)}_i\right)\right),
    \label{eq:adapter_basic}
\end{equation}
%
Then, as shown in Fig.~\ref{fig:framework}(a), the `augmented/adapted' representation $\hat{\mathbf{x}}^{(l)}_i$ 
can be obtained 
by integrating the pre-trained GNN's representation ${\mathbf{y}}^{(l)}_i$
and adapter's output $\mathbf{z}^{(l)}_i$ (Eq.~(\ref{eq:adapter_basic})) together as 
\begin{equation}
 {\mathbf{\hat{x}}}^{(l)}_i =    {\mathrm{BN}(\mathbf{y}}^{(l)}_i) +\mathbf{z}^{(l)}_i.
\end{equation}
To preserve the task-agnostic knowledge in pre-trained model \(f_{\Omega}\), one can freeze the pre-trained model and optimize the above lightweight adapter module (Eq.~(\ref{eq:adapter_basic})) parameterized by $\Theta^{(l)}=\{\mathbf{W}^{(l)}_{\mathrm{down}},\mathbf{W}^{(l)}_{\mathrm{up}}\}$, as well as a task-specific classifier based on downstream
labeled graph data. 

\section{Methodology}
In this section, 
we propose our Uncertainty-aware Adapter (UAdapterGNN) framework to enhance the robustness and generalization of pre-trained GNNs on the downstream tasks. 
The whole fine-tuning process is illustrated in Fig.~\ref{fig:framework}(b). It mainly comprises three main components: 1) message passing via a frozen pre-trained GNN backbone, 2) uncertainty representation via Gaussian random adapter and 3) feature augmentation  via re-parameterized operator. 
Below, we introduce them in detail.

\subsection{Message Passing via  Frozen Pre-trained GNN}

Given a pre-trained GNN encoder $f_{\Omega}$ with frozen parameters $\Omega$ and input graph $\mathcal{G} = (\mathcal{V}, \mathcal{E}, \mathbf{X}, \mathbf{A})$, similar to the above message passing Eq.(\ref{eq: massage_passing}) in standard AdapterGNN~\cite{adaptergnn}, the message passing at $l$-th layer can be generally formulated as follows, 
\begin{equation}
{\mathbf{y}}_i^{(l)} =\mathrm{MLP} \Big( \bigoplus_{j \in \mathcal{N}(i)} m^{(l)} \big( {\mathbf{x}}_i^{(l)}, {\mathbf{x}}_j^{(l)}, {e}_{ji} \big) \Big)
\end{equation}
%
where 
$\mathcal{N}(i)$ indicates the neighborhood of node $v_i$ and $\bigoplus$ denotes a permutation-invariant aggregation operator (typically summation or mean operator). $m^{(l)}$ 
represents the message function. All parameters in $\Omega$  remain frozen during the fine-tuning process. 

\subsection{Uncertain Modeling via Gaussian Random Adapter}
%
Unlike traditional AdapterGNN that outputs 
task-related node representation  $\mathbf{z}^{(l)}_i$ (Eq.~(\ref{eq:adapter_basic})) via a lightweight bottleneck layer, 
our UAdapterGNN models task-related node representation $\mathbf{z}^{(l)}_i$ as a Gaussian distribution that allows adaptive adjustment to noise-induced variations. 
To obtain $\mathbf{z}^{(l)}_i$, 
we adopt two parameterized modules to estimate the mean $\mathbf{\mu}^{(l)}_i$ and variance $\mathbf{\sigma}^{(l)}_i$, respectively. 
To be specific,  
for the input ${\mathbf{x}}_i^{(l)}$, we obtain the \emph{mean} vector via 
a bottleneck layer as, 
\begin{equation}
    \mu^{(l)}_i =  \mathrm{BN}\left( \mathbf{W}^{(l)}_{\text{up}} \cdot \mathrm{ReLU}( \mathbf{W}^{(l)}_{\text{down}} \cdot {\mathbf{x}}_i^{(l)} ) \right)
\end{equation}
where $\mathbf{W}^{(l)}_{\text{down}}$ and $\mathbf{W}^{(l)}_{\text{up}}$ denote the layer-specific trainable parameters. Similarly, the \emph{variance} vector can be computed via another bottleneck layer as
\begin{equation}
    {\sigma}^{(l)}_i = \mathrm{BN}\left( \mathbf{\tilde{W}}^{(l)}_{\text{up}} \cdot \mathrm{ReLU}\big( \mathbf{\tilde{W}}^{(l)}_{\text{down}} \cdot {\mathbf{x}}_i^{(l)} \big) \right)
\end{equation}
where $\mathbf{\tilde{W}}^{(l)}_{\text{down}}$ and $\mathbf{\tilde{W}}^{(l)}_{\text{up}}$ are learnable parameters. 
Based on learned ${\mu}^{(l)}_i$ and ${\sigma}^{(l)}_i$, we can obtain the Gaussian distribution representation for node $i$
as
\begin{equation}
\mathbf{z}^{(l)}_i \sim \mathcal{N}\big(\mathbf{\mu}^{(l)}_i, {\mathrm{diag}(\mathbf{\sigma}}^{(l)}_i)\big)
\end{equation}

%

\subsection{Re-parameterized Operator}

After obtaining the above Gaussian random adapter representation $\mathbf{z}^{(l)}_i$, we need to integrate it 
with the original pre-trained GNN layer's output $\mathbf{y}^{(l)}_i$ for feature augmentation. 
One simple way is to conduct sampling based on the learned Gaussian distribution  $\mathcal{N}(\mu^{(l)}_i,\mathrm{diag}(\sigma^{(l)}_i))$. 
However, direct sampling from the distribution leads to non-differentiable operation that prevents gradient-based optimization. 
Therefore, we employ a re-parameterization trick~\cite{vae} to enable end-to-end training.
Specifically, we first sample a random noise vector $\epsilon$ from the standard normal distribution and then compute the equivalent sampled representation as follows, 
\begin{equation}
    \mathbf{{z}}^{(l)}_i = \mu^{(l)}_i+\epsilon \odot \sigma^{(l)}_i,\quad \epsilon \sim \mathcal{N}(\textbf{0},\;\textbf{I}),
\end{equation}
After that, 
we introduce an adaptive integration mechanism based on learnable scaling factor to dynamically balance the pre-trained GNN layer's output and the uncertainty-aware adaptation $\mathbf{{z}}^{(l)}_i$. Therefore, the final node feature augmentation is obtained as follows,  
\begin{equation}
  \mathbf{\hat{x}}^{(l)}_i 
  = {\mathrm{BN}(\mathbf{y}}^{(l)}_i)
   \;+\; s^{(l)} \cdot \mathbf{z}^{(l)}_i 
\label{eq:final_integration}
\end{equation}
The scaling factor $s^{(l)}$ is defined as the trainable parameter during the fine-tuning process. 
Finally, the augmentation representation 
${\mathbf{\hat{x}}}^{(l)}_i$ is passed to the next GNN layer and iteratively refined by the next uncertainty-aware adapter. 

\textbf{Remark.} Figure~\ref{fig:framework}(b) shows the architecture of our UAdapterGNN module. Note that, similar to traditional AdapterGNN~\cite{adaptergnn} (Fig.~\ref{fig:framework}(a)), it offers a general plug-and-play adapter architecture which can be integrated with
any pre-trained GNN models to enable robust learning on the
downstream learning tasks, as validated in Experiments below. 


\begin{table*}[!htbp]
\centering
\caption{Test ROC-AUC (\%)  performence on molecular prediction benchmarks using various pre-training strategies and tuning methods.}
\label{tab:benchmark}
\renewcommand{\arraystretch}{1.1} 
\definecolor{lightgray}{gray}{0.5} 
\resizebox{\linewidth}{!}{
\begin{tabular}{c c c c c c c c c c | c}
\toprule
\multicolumn{1}{c}{Tuning} & \multicolumn{1}{c}{Pre-training} & \multicolumn{9}{c}{Dataset} \\
 Method & Method & BACE & BBBP & ClinTox & HIV & SIDER & Tox21 & MUV & ToxCast & Avg. \\
\midrule

\multirow{5}{*}{\begin{tabular}[c]{@{}c@{}}Full Fine-tune\\(100\%)\end{tabular}} 
& EdgePred & 79.9\(_{\textcolor{lightgray}{\pm0.9}}\) & 67.3\(_{\textcolor{lightgray}{\pm2.4}}\) & 64.1\(_{\textcolor{lightgray}{\pm3.7}}\) & 76.3\(_{\textcolor{lightgray}{\pm1.0}}\) & 60.4\(_{\textcolor{lightgray}{\pm0.7}}\) & \textbf{76.0\(_{\textcolor{lightgray}{\pm0.6}}\)} & 74.1\(_{\textcolor{lightgray}{\pm2.1}}\) & 64.1\(_{\textcolor{lightgray}{\pm0.6}}\) & 70.3 \\
& ContextPred & 79.6\(_{\textcolor{lightgray}{\pm1.2}}\) & 68.0\(_{\textcolor{lightgray}{\pm2.0}}\) & 65.9\(_{\textcolor{lightgray}{\pm3.8}}\) & \textbf{77.3\(_{\textcolor{lightgray}{\pm1.0}}\)} & 60.9\(_{\textcolor{lightgray}{\pm0.6}}\) & \textbf{75.7\(_{\textcolor{lightgray}{\pm0.7}}\)} & 75.8\(_{\textcolor{lightgray}{\pm1.7}}\) & \textbf{63.9\(_{\textcolor{lightgray}{\pm0.6}}\)} & 70.9 \\
& AttrMasking & 79.3\(_{\textcolor{lightgray}{\pm1.6}}\) & 64.3\(_{\textcolor{lightgray}{\pm2.8}}\) & 71.8\(_{\textcolor{lightgray}{\pm4.1}}\) & \textbf{77.2\(_{\textcolor{lightgray}{\pm1.1}}\)} & 61.0\(_{\textcolor{lightgray}{\pm0.7}}\) & \textbf{76.7\(_{\textcolor{lightgray}{\pm0.4}}\)} & 74.7\(_{\textcolor{lightgray}{\pm1.4}}\) & \textbf{64.2\(_{\textcolor{lightgray}{\pm0.5}}\)} & 71.1 \\
& GraphCL & 74.6\(_{\textcolor{lightgray}{\pm2.2}}\) & 68.6\(_{\textcolor{lightgray}{\pm2.3}}\) & 69.8\(_{\textcolor{lightgray}{\pm7.2}}\) & 78.5\(_{\textcolor{lightgray}{\pm1.2}}\) & 59.6\(_{\textcolor{lightgray}{\pm0.7}}\) & 74.4\(_{\textcolor{lightgray}{\pm0.5}}\) & 73.7\(_{\textcolor{lightgray}{\pm2.7}}\) & 62.9\(_{\textcolor{lightgray}{\pm0.4}}\) & 70.3 \\
& SimGRACE & 74.7\(_{\textcolor{lightgray}{\pm1.0}}\) & 69.0\(_{\textcolor{lightgray}{\pm1.0}}\) & 59.9\(_{\textcolor{lightgray}{\pm2.3}}\) & 74.6\(_{\textcolor{lightgray}{\pm1.2}}\) & 59.1\(_{\textcolor{lightgray}{\pm0.6}}\) & 73.9\(_{\textcolor{lightgray}{\pm0.4}}\) & 71.0\(_{\textcolor{lightgray}{\pm1.9}}\) & 61.8\(_{\textcolor{lightgray}{\pm0.4}}\) & 68.0 \\
\midrule

\multirow{5}{*}{\begin{tabular}[c]{@{}c@{}}Adapter\\(5.2\%)\end{tabular}} 
& EdgePred & 78.5\(_{\textcolor{lightgray}{\pm1.7}}\) & 65.9\(_{\textcolor{lightgray}{\pm2.8}}\) & 66.6\(_{\textcolor{lightgray}{\pm5.4}}\) & 73.5\(_{\textcolor{lightgray}{\pm0.2}}\) & 60.9\(_{\textcolor{lightgray}{\pm1.3}}\) & 75.4\(_{\textcolor{lightgray}{\pm0.5}}\) & 73.0\(_{\textcolor{lightgray}{\pm1.0}}\) & 63.0\(_{\textcolor{lightgray}{\pm0.7}}\) & 69.6 \\
& ContextPred & 75.0\(_{\textcolor{lightgray}{\pm3.3}}\) & 68.2\(_{\textcolor{lightgray}{\pm3.0}}\) & 57.6\(_{\textcolor{lightgray}{\pm3.6}}\) & 75.4\(_{\textcolor{lightgray}{\pm0.6}}\) & \textbf{62.4\(_{\textcolor{lightgray}{\pm1.2}}\)} & 74.7\(_{\textcolor{lightgray}{\pm0.7}}\) & 73.3\(_{\textcolor{lightgray}{\pm0.8}}\) & 62.2\(_{\textcolor{lightgray}{\pm0.4}}\) & 68.6 \\
& AttrMasking & 76.1\(_{\textcolor{lightgray}{\pm1.4}}\) & 68.7\(_{\textcolor{lightgray}{\pm1.7}}\) & 65.8\(_{\textcolor{lightgray}{\pm4.4}}\) & 75.6\(_{\textcolor{lightgray}{\pm0.7}}\) & 59.8\(_{\textcolor{lightgray}{\pm1.7}}\) & 74.4\(_{\textcolor{lightgray}{\pm0.9}}\) & 75.8\(_{\textcolor{lightgray}{\pm2.4}}\) & 62.6\(_{\textcolor{lightgray}{\pm0.8}}\) & 69.9 \\
& GraphCL & 72.5\(_{\textcolor{lightgray}{\pm3.0}}\) & 69.3\(_{\textcolor{lightgray}{\pm0.6}}\) & 67.3\(_{\textcolor{lightgray}{\pm7.4}}\) & 75.0\(_{\textcolor{lightgray}{\pm0.4}}\) & \textbf{59.7\(_{\textcolor{lightgray}{\pm1.2}}\)} & 74.7\(_{\textcolor{lightgray}{\pm0.4}}\) & 72.9\(_{\textcolor{lightgray}{\pm1.7}}\) & 62.9\(_{\textcolor{lightgray}{\pm0.4}}\) & 69.3 \\
& SimGRACE & 73.4\(_{\textcolor{lightgray}{\pm1.1}}\) & 64.8\(_{\textcolor{lightgray}{\pm0.7}}\) & 63.5\(_{\textcolor{lightgray}{\pm4.4}}\) & 73.9\(_{\textcolor{lightgray}{\pm1.0}}\) & \textbf{59.9\(_{\textcolor{lightgray}{\pm0.9}}\)} & 73.1\(_{\textcolor{lightgray}{\pm0.9}}\) & 70.1\(_{\textcolor{lightgray}{\pm4.6}}\) & 61.7\(_{\textcolor{lightgray}{\pm0.8}}\) & 67.6 \\
\midrule

\multirow{5}{*}{\begin{tabular}[c]{@{}c@{}}LoRA\\(5.0\%)\end{tabular}} 
& EdgePred & \textbf{81.0\(_{\textcolor{lightgray}{\pm0.8}}\)} & 64.8\(_{\textcolor{lightgray}{\pm1.6}}\) & 67.7\(_{\textcolor{lightgray}{\pm1.2}}\) & 74.8\(_{\textcolor{lightgray}{\pm1.2}}\) & 60.8\(_{\textcolor{lightgray}{\pm1.1}}\) & 74.6\(_{\textcolor{lightgray}{\pm0.4}}\) & 75.0\(_{\textcolor{lightgray}{\pm1.5}}\) & 62.2\(_{\textcolor{lightgray}{\pm1.0}}\) & 70.1 \\
& ContextPred & 78.5\(_{\textcolor{lightgray}{\pm1.1}}\) & 65.3\(_{\textcolor{lightgray}{\pm2.4}}\) & 61.3\(_{\textcolor{lightgray}{\pm1.9}}\) & 74.7\(_{\textcolor{lightgray}{\pm1.6}}\) & 60.8\(_{\textcolor{lightgray}{\pm0.4}}\) & 72.9\(_{\textcolor{lightgray}{\pm0.4}}\) & 75.4\(_{\textcolor{lightgray}{\pm0.9}}\) & 63.4\(_{\textcolor{lightgray}{\pm0.2}}\) & 69.0 \\
& AttrMasking & 79.8\(_{\textcolor{lightgray}{\pm0.7}}\) & 64.2\(_{\textcolor{lightgray}{\pm1.1}}\) & 70.1\(_{\textcolor{lightgray}{\pm2.9}}\) & 76.1\(_{\textcolor{lightgray}{\pm1.4}}\) & 59.7\(_{\textcolor{lightgray}{\pm0.5}}\) & 74.6\(_{\textcolor{lightgray}{\pm0.5}}\) & 76.6\(_{\textcolor{lightgray}{\pm1.6}}\) & 61.7\(_{\textcolor{lightgray}{\pm0.4}}\) & 70.4 \\
& GraphCL & 75.1\(_{\textcolor{lightgray}{\pm0.7}}\) & 67.8\(_{\textcolor{lightgray}{\pm1.1}}\) & 65.1\(_{\textcolor{lightgray}{\pm3.5}}\) & \textbf{78.9\(_{\textcolor{lightgray}{\pm0.6}}\)} & 57.6\(_{\textcolor{lightgray}{\pm0.7}}\) & 73.9\(_{\textcolor{lightgray}{\pm0.9}}\) & 72.8\(_{\textcolor{lightgray}{\pm1.2}}\) & 62.7\(_{\textcolor{lightgray}{\pm0.6}}\) & 69.2 \\
& SimGRACE & 73.2\(_{\textcolor{lightgray}{\pm1.0}}\) & 67.5\(_{\textcolor{lightgray}{\pm0.4}}\) & 60.7\(_{\textcolor{lightgray}{\pm0.4}}\) & 74.1\(_{\textcolor{lightgray}{\pm0.5}}\) & 57.6\(_{\textcolor{lightgray}{\pm2.6}}\) & 72.2\(_{\textcolor{lightgray}{\pm0.2}}\) & 67.9\(_{\textcolor{lightgray}{\pm0.9}}\) & 61.8\(_{\textcolor{lightgray}{\pm0.2}}\) & 66.9 \\
\midrule
\midrule
\multirow{4}{*}{\begin{tabular}[c]{@{}c@{}}GPF\\(0.1\%)\end{tabular}} 
& EdgePred & 68.0\(_{\textcolor{lightgray}{\pm0.4}}\) & 55.9\(_{\textcolor{lightgray}{\pm0.2}}\) & 50.8\(_{\textcolor{lightgray}{\pm0.1}}\) & 66.0\(_{\textcolor{lightgray}{\pm0.7}}\) & 51.5\(_{\textcolor{lightgray}{\pm0.7}}\) & 63.1\(_{\textcolor{lightgray}{\pm0.5}}\) & 63.1\(_{\textcolor{lightgray}{\pm0.1}}\) & 55.7\(_{\textcolor{lightgray}{\pm0.5}}\) & 59.3 \\
& ContextPred & 58.7\(_{\textcolor{lightgray}{\pm0.6}}\) & 58.6\(_{\textcolor{lightgray}{\pm0.6}}\) & 39.8\(_{\textcolor{lightgray}{\pm0.8}}\) & 68.0\(_{\textcolor{lightgray}{\pm0.4}}\) & 59.4\(_{\textcolor{lightgray}{\pm0.2}}\) & 67.8\(_{\textcolor{lightgray}{\pm0.9}}\) & 71.8\(_{\textcolor{lightgray}{\pm0.8}}\) & 58.8\(_{\textcolor{lightgray}{\pm0.5}}\) & 60.4 \\
& AttrMasking & 61.7\(_{\textcolor{lightgray}{\pm0.3}}\) & 54.3\(_{\textcolor{lightgray}{\pm0.3}}\) & 56.4\(_{\textcolor{lightgray}{\pm0.2}}\) & 64.0\(_{\textcolor{lightgray}{\pm0.2}}\) & 52.0\(_{\textcolor{lightgray}{\pm0.2}}\) & 69.2\(_{\textcolor{lightgray}{\pm0.3}}\) & 62.9\(_{\textcolor{lightgray}{\pm0.9}}\) & 58.1\(_{\textcolor{lightgray}{\pm0.3}}\) & 59.8 \\
& GraphCL & 71.5\(_{\textcolor{lightgray}{\pm0.6}}\) & 63.7\(_{\textcolor{lightgray}{\pm0.4}}\) & 64.5\(_{\textcolor{lightgray}{\pm0.6}}\) & 70.3\(_{\textcolor{lightgray}{\pm0.5}}\) & 55.3\(_{\textcolor{lightgray}{\pm0.6}}\) & 65.5\(_{\textcolor{lightgray}{\pm0.5}}\) & 70.1\(_{\textcolor{lightgray}{\pm0.7}}\) & 58.5\(_{\textcolor{lightgray}{\pm0.5}}\) & 64.9 \\
\midrule
MolCPT (40.0\%) & GraphCL & 74.1\(_{\textcolor{lightgray}{\pm0.5}}\) & 60.5\(_{\textcolor{lightgray}{\pm0.8}}\) & 73.4\(_{\textcolor{lightgray}{\pm0.8}}\) & 64.5\(_{\textcolor{lightgray}{\pm0.8}}\) & 55.9\(_{\textcolor{lightgray}{\pm0.3}}\) & 67.4\(_{\textcolor{lightgray}{\pm0.7}}\) & 65.7\(_{\textcolor{lightgray}{\pm2.2}}\) & 57.9\(_{\textcolor{lightgray}{\pm0.3}}\) & 64.9 \\
\midrule
\midrule

\multirow{5}{*}{\begin{tabular}[c]{@{}c@{}}AdapterGNN\\(5.2\%)\end{tabular}} 
& EdgePred & 79.0\(_{\textcolor{lightgray}{\pm1.5}}\) & 69.7\(_{\textcolor{lightgray}{\pm1.4}}\) & 67.7\(_{\textcolor{lightgray}{\pm3.0}}\) & 76.4\(_{\textcolor{lightgray}{\pm0.7}}\) & 61.2\(_{\textcolor{lightgray}{\pm0.9}}\) & 75.9\(_{\textcolor{lightgray}{\pm0.9}}\) & 75.8\(_{\textcolor{lightgray}{\pm2.1}}\) & \textbf{64.2\(_{\textcolor{lightgray}{\pm0.5}}\)} & 71.2 \\
& ContextPred & 78.7\(_{\textcolor{lightgray}{\pm2.0}}\) & 68.2\(_{\textcolor{lightgray}{\pm2.9}}\) & 68.7\(_{\textcolor{lightgray}{\pm5.3}}\) & 76.1\(_{\textcolor{lightgray}{\pm0.5}}\) & 61.1\(_{\textcolor{lightgray}{\pm1.0}}\) & 75.4\(_{\textcolor{lightgray}{\pm0.6}}\) & 76.3\(_{\textcolor{lightgray}{\pm1.0}}\) & 63.2\(_{\textcolor{lightgray}{\pm0.3}}\) & 71.0 \\
& AttrMasking & 79.7\(_{\textcolor{lightgray}{\pm1.3}}\) & 67.5\(_{\textcolor{lightgray}{\pm2.2}}\) & \textbf{78.3\(_{\textcolor{lightgray}{\pm2.6}}\)} & 76.7\(_{\textcolor{lightgray}{\pm1.2}}\) & 61.3\(_{\textcolor{lightgray}{\pm1.1}}\) & 76.6\(_{\textcolor{lightgray}{\pm0.5}}\) & 78.4\(_{\textcolor{lightgray}{\pm0.7}}\) & 63.6\(_{\textcolor{lightgray}{\pm0.5}}\) & 72.8 \\
& GraphCL & 76.1\(_{\textcolor{lightgray}{\pm2.2}}\) & 67.8\(_{\textcolor{lightgray}{\pm1.4}}\) & 72.0\(_{\textcolor{lightgray}{\pm3.8}}\) & 77.8\(_{\textcolor{lightgray}{\pm1.3}}\) & 59.6\(_{\textcolor{lightgray}{\pm1.3}}\) & 74.9\(_{\textcolor{lightgray}{\pm0.9}}\) & 75.1\(_{\textcolor{lightgray}{\pm2.1}}\) & \textbf{63.1\(_{\textcolor{lightgray}{\pm0.4}}\)} & 70.7 \\
& SimGRACE & 77.7\(_{\textcolor{lightgray}{\pm1.7}}\) & 68.1\(_{\textcolor{lightgray}{\pm1.3}}\) & 73.9\(_{\textcolor{lightgray}{\pm7.0}}\) & 75.1\(_{\textcolor{lightgray}{\pm1.2}}\) & 58.9\(_{\textcolor{lightgray}{\pm0.9}}\) & 74.4\(_{\textcolor{lightgray}{\pm0.6}}\) & 71.8\(_{\textcolor{lightgray}{\pm1.4}}\) & 62.6\(_{\textcolor{lightgray}{\pm0.6}}\) & 70.3 \\
\midrule

\multirow{5}{*}{\begin{tabular}[c]{@{}c@{}}\textbf{UAdapterGNN}\\(5.2\%)\end{tabular}} 
& EdgePred & 80.7\(_{\textcolor{lightgray}{\pm1.0}}\) & \textbf{71.1\(_{\textcolor{lightgray}{\pm1.0}}\)} & \textbf{70.9\(_{\textcolor{lightgray}{\pm3.6}}\)} & \textbf{76.6\(_{\textcolor{lightgray}{\pm0.7}}\)} & \textbf{62.2\(_{\textcolor{lightgray}{\pm0.9}}\)} & 75.6\(_{\textcolor{lightgray}{\pm0.5}}\) & \textbf{78.7\(_{\textcolor{lightgray}{\pm0.7}}\)} & \textbf{64.2\(_{\textcolor{lightgray}{\pm0.5}}\)} & \textbf{72.5} \\
& ContextPred & \textbf{80.3\(_{\textcolor{lightgray}{\pm0.6}}\)} & \textbf{69.6\(_{\textcolor{lightgray}{\pm1.8}}\)} & \textbf{74.3\(_{\textcolor{lightgray}{\pm1.7}}\)} & 76.8\(_{\textcolor{lightgray}{\pm0.8}}\) & 62.3\(_{\textcolor{lightgray}{\pm0.8}}\) & \textbf{75.7\(_{\textcolor{lightgray}{\pm0.5}}\)} & \textbf{79.4\(_{\textcolor{lightgray}{\pm1.3}}\)} & \textbf{63.9\(_{\textcolor{lightgray}{\pm0.4}}\)} & \textbf{72.8} \\
& AttrMasking & \textbf{80.4\(_{\textcolor{lightgray}{\pm0.9}}\)} & \textbf{69.6\(_{\textcolor{lightgray}{\pm2.7}}\)} & 77.2\(_{\textcolor{lightgray}{\pm1.2}}\) & 76.8\(_{\textcolor{lightgray}{\pm0.7}}\) & \textbf{62.8\(_{\textcolor{lightgray}{\pm0.8}}\)} & \textbf{76.7\(_{\textcolor{lightgray}{\pm0.5}}\)} & \textbf{80.1\(_{\textcolor{lightgray}{\pm1.6}}\)} & 63.9\(_{\textcolor{lightgray}{\pm0.5}}\) & \textbf{73.4} \\
& GraphCL & \textbf{77.3\(_{\textcolor{lightgray}{\pm2.9}}\)} & \textbf{70.2\(_{\textcolor{lightgray}{\pm0.7}}\)} & \textbf{74.7\(_{\textcolor{lightgray}{\pm2.4}}\)} & 77.3\(_{\textcolor{lightgray}{\pm0.5}}\) & \textbf{59.7\(_{\textcolor{lightgray}{\pm0.7}}\)} & \textbf{75.5\(_{\textcolor{lightgray}{\pm0.5}}\)} & \textbf{76.9\(_{\textcolor{lightgray}{\pm1.6}}\)} & \textbf{62.9\(_{\textcolor{lightgray}{\pm0.4}}\)} & \textbf{71.8} \\
& SimGRACE & \textbf{80.4\(_{\textcolor{lightgray}{\pm1.7}}\)} & \textbf{70.6\(_{\textcolor{lightgray}{\pm0.6}}\)} & \textbf{76.4\(_{\textcolor{lightgray}{\pm1.2}}\)} & \textbf{75.2\(_{\textcolor{lightgray}{\pm0.6}}\)} & 59.6\(_{\textcolor{lightgray}{\pm0.5}}\) & \textbf{74.9\(_{\textcolor{lightgray}{\pm0.4}}\)} & \textbf{74.5\(_{\textcolor{lightgray}{\pm2.3}}\)} & \textbf{63.1\(_{\textcolor{lightgray}{\pm0.3}}\)} & \textbf{71.8} \\
\bottomrule
\end{tabular}}
\end{table*}

\section{Experiments}
In this section, we conduct experiments to assess the performance and benefits of the proposed UAdapterGNN.

\subsection{Experimental Setup}

\textbf{Datasets.} 
In experiments, we utilize the chemistry datasets used in previous works~\cite{hu2019strategies,adaptergnn} for pre-training and downstream tasks. The pre-training dataset includes 2 million unlabeled molecules sampled from the ZINC15 database~\cite{sterling2015zinc}, along with 456K labeled molecules from the preprocessed ChEMBL dataset~\cite{mayr2018large, gaulton2012chembl}. For downstream tasks, we evaluate the proposed  UAdapterGNN on eight molecular classification datasets from MoleculeNet~\cite{wu2018moleculenet}. 


\textbf{Pre-training Strategies. } 
We adopt five widely used strategies to pre-train the model, which are listed as below:
\begin{itemize}
\item \textit{AttrMasking~\cite{hu2019strategies}} aims to mask node or edge attributes and then train GNNs by predicting these attributes.
\item \textit{ContextPred~\cite{hu2019strategies}} proposes to perform node or subgraph context prediction for GNNs pre-training.
\item \textit{EdgePred~\cite{GraphSAGE}} predicts the existence of edge between node pairs by structure reconstruction task.  
\item  \textit{GraphCL~\cite{you2020graph}} contrasts embeddings between different augmented graph views.
\item  \textit{SimGRACE~\cite{xia2022simgrace}} contrasts embeddings between original and perturbed GNN encoders without data augmentation.
\end{itemize}

\textbf{Implementation Details. }
We adopt 5-layer Graph Isomorphism Network (GIN)~\cite{xu2018powerful} as the GNN backbone with the hidden dimension of $300$. 
In our experiments,  the bottleneck dimension of adapter module is tuned from $\{15, 20, 30\}$ to balance efficiency and representation capacity. We select the learning rate from $\{0.001, 0.005\}$ and then train the model using Adam optimizer~\cite{Kingma2014AdamAM} with epochs of $100$ and batch size of $32$. The number of sampling instances is varied from $\{1, 3, 5, 7\}$. The scaling coefficient with initialization of $0.01$ is dynamically learned to balance task-specific adaptation and pre-trained knowledge retention. All experiments are run five times with different random seeds.

\subsection{Comparison with Related Works}
\textbf{Comparison Methods. }
To validate the effectiveness of our UAdapterGNN, we compare it with some graph fine-tuning methods as follows. First, we consider the full fine-tuning method, which updates all parameters of the pre-trained GNN. Then, we select several parameter-efficient fine-tuning (PEFT) methods adapted to graph-structured data for comparison: (1) Adapter method: Adapter~\cite{adapter} inserts lightweight adapter modules into the model layers and AdapterGNN~\cite{adaptergnn} incorporates dual parallel adapters in each GNN layer. (2) LoRA method: LoRA~\cite{hu2022lora} optimizes a subset of parameters via low-rank decomposition. (3) Prompting method: GPF~\cite{gpf} appends learnable features to input embeddings with minimal parameter overhead and MolCPT~\cite{molcpt} injects molecular motif  prompts to augment molecular graphs in latent space. For fairness, all comparison baselines utilize identical pre-trained weights from established strategies  which are reproduced by using their provided implementations.

\begin{table*}[!htbp]
\centering
\normalsize
\caption{Robustness analysis with random structural noise injection on Tox21 and ToxCast datasets.}
\label{tab:robustness}
\definecolor{lightgray}{gray}{0.5} 
\resizebox{0.85\linewidth}{!}{
\setlength{\tabcolsep}{3pt}  
\renewcommand{\arraystretch}{1.25}  
\begin{tabular}{@{} c| c| *{5}{c} |*{5}{c} @{}}  
\toprule
& & \multicolumn{5}{c|}{Edge Deletion Level} 
& \multicolumn{5}{c}{Edge Addition Level} \\
\midrule
Dataset & Methods & 0\% & 20\% & 40\% & 60\% & 80\% & 0\% & 20\% & 40\% & 60\% & 80\% \\
\midrule
\multirow{3}{*}{Tox21} 
& Full Fine-tune & 76.25 & 71.07 & 67.24 & 65.68 & 65.26 & 76.25 & 72.41 & 68.26 & 66.69 & 63.00 \\
& AdapterGNN & 76.41 & 71.41 & 67.59 & 66.15 & 65.20 & 76.41 & 72.60 & 68.81 & 66.62 & 64.00  \\
& UAdapterGNN(Ours)  & 76.58 & 71.93 & 68.45 & 66.90 & 67.21 & 76.58 & 72.98 & 69.54 & 68.08 & 66.61 \\
\midrule
\multirow{3}{*}{Toxcast}
& Full Fine-tune & 62.96  & 60.88  & 58.89  & 56.82  & 55.49  & 62.96  & 61.03  & 58.02  & 56.70  & 55.43  \\
& AdapterGNN & 63.32  & 61.55  & 59.25  & 57.11  & 55.66  & 63.32  & 60.75  & 57.71  & 56.72  & 55.61  \\
& UAdapterGNN(Ours) & 63.88  & 61.64  & 60.02  & 57.92  & 57.21  & 63.88  & 61.38  & 58.36  & 57.67  & 56.78  \\
\bottomrule
\end{tabular}}
\end{table*}

\begin{figure*}[!htbp]
    \centering
    \includegraphics[width=0.8\linewidth]{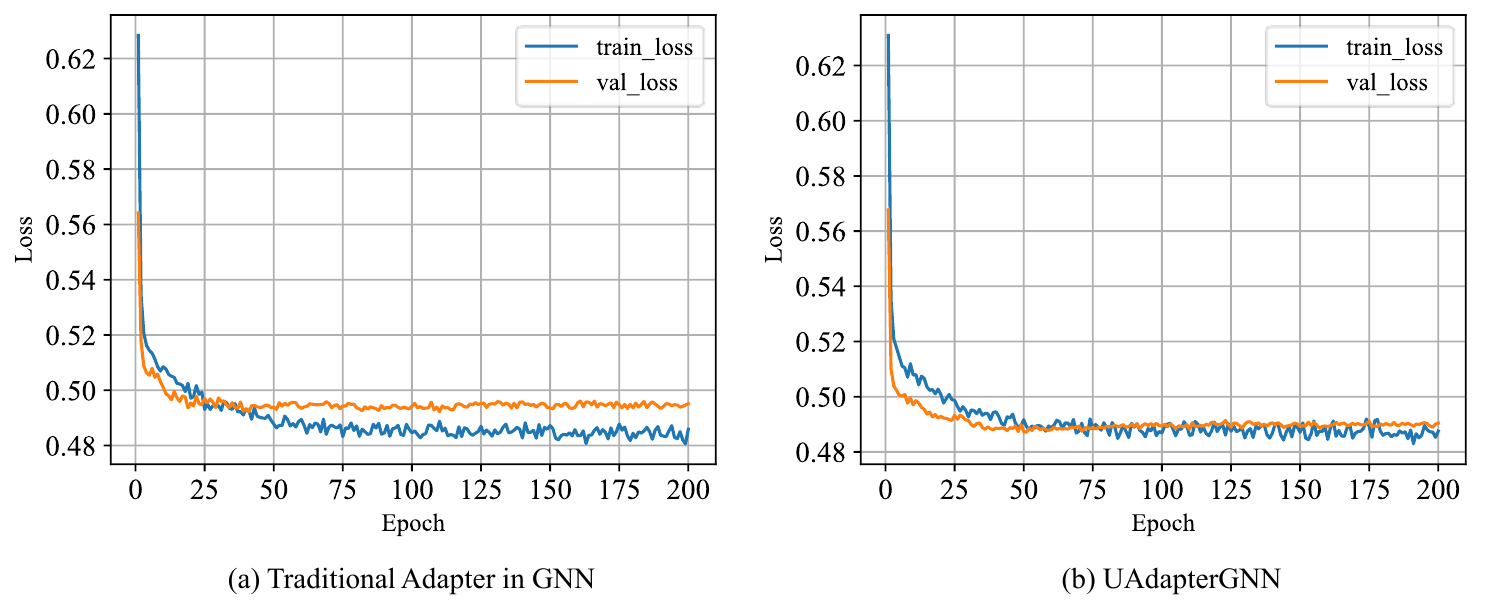}
    \caption{Generalization result on SIDER dataset. As shown in the figure, the variant without the Gaussion random adapter exhibits a notably larger gap, demonstrating the effectiveness of our uncertainty-aware adapter in enhancing the model’s generalization performance.}
    \label{fig:sider_generalization}
\end{figure*}

\textbf{Comparison Results. }
Table~\ref{tab:benchmark} reports the comparison results of all methods on eight datasets. Here, we can observe that: 
(1) Compared to conventional fine-tuning, parameter-efficient methods obtain competitive performance with only a small fraction of trainable parameters, demonstrating the effectiveness and efficiency of PEFT in enhancing model adaptability and downstream performance.
(2) Compared to LoRA~\cite{hu2022lora} and prompt based PEFT methods, our UAdapterGNN achieves better performance on almost all datasets, which indicates the advantages of adapter tuning in model knowledge transferring.
(3) Moreover, compared to other adapter-style methods including {Adapter}~\cite{adapter}, {LoRA}~\cite{hu2022lora}, {AdapterGNN}~\cite{adaptergnn} with comparable parameter budgets, UAdapterGNN consistently attains the best results on all datasets, achieving an average ROC-AUC of 72.46\%, corresponding to an approximate 1.3\% –3.5\% improvements. This clearly validates the effectiveness of introducing uncertainty modeling via Gaussian random adapter in graph adaptation fine-tuning. 

\begin{table}[!htbp]
\centering
\caption{Comparison of our learnable scaling and fixed scaling.}
\label{tab:scaling_ablation}
\resizebox{\linewidth}{!}{
\setlength{\tabcolsep}{3pt}
\renewcommand{\arraystretch}{1.25}  
\definecolor{lightgray}{gray}{0.5} 
\begin{tabular}{c c c c c c c c}
\toprule
 & BACE & BBBP & ClinTox & SIDER & Tox21 & ToxCast & Avg. \\ \midrule
0.01 & 79.6\(_{\textcolor{lightgray}{\pm0.8}}\) & 67.9\(_{\textcolor{lightgray}{\pm2.4}}\) & 75.3\(_{\textcolor{lightgray}{\pm1.0}}\) & 59.0\(_{\textcolor{lightgray}{\pm0.6}}\) & 74.7\(_{\textcolor{lightgray}{\pm0.5}}\) & 62.4\(_{\textcolor{lightgray}{\pm0.5}}\) & 69.8 \\
0.1 & 79.8\(_{\textcolor{lightgray}{\pm1.6}}\) & 69.4\(_{\textcolor{lightgray}{\pm1.9}}\) & 74.8\(_{\textcolor{lightgray}{\pm1.7}}\) & 62.1\(_{\textcolor{lightgray}{\pm0.5}}\) & 75.4\(_{\textcolor{lightgray}{\pm0.4}}\) & 63.1\(_{\textcolor{lightgray}{\pm1.0}}\) & 70.8 \\
0.5 & 79.3\(_{\textcolor{lightgray}{\pm1.7}}\) & 62.0\(_{\textcolor{lightgray}{\pm3.3}}\) & 72.9\(_{\textcolor{lightgray}{\pm2.3}}\) & 61.8\(_{\textcolor{lightgray}{\pm0.7}}\) & 75.4\(_{\textcolor{lightgray}{\pm0.5}}\) & 63.5\(_{\textcolor{lightgray}{\pm0.3}}\)& 69.2 \\
1 & 79.4\(_{\textcolor{lightgray}{\pm1.0}}\) & 66.6\(_{\textcolor{lightgray}{\pm1.5}}\) & 68.7\(_{\textcolor{lightgray}{\pm5.3}}\) & 61.0\(_{\textcolor{lightgray}{\pm0.4}}\) & 75.2\(_{\textcolor{lightgray}{\pm0.4}}\) & 63.4\(_{\textcolor{lightgray}{\pm0.4}}\) & 69.1 \\
5 & 68.3\(_{\textcolor{lightgray}{\pm9.5}}\) & 62.7\(_{\textcolor{lightgray}{\pm1.0}}\) & 49.6\(_{\textcolor{lightgray}{\pm4.3}}\) & 59.8\(_{\textcolor{lightgray}{\pm1.1}}\) & 73.5\(_{\textcolor{lightgray}{\pm0.7}}\) & 62.3\(_{\textcolor{lightgray}{\pm0.6}}\) & 62.7 \\
\textbf{Ours} & \textbf{80.4\(_{\textcolor{lightgray}{\pm0.9}}\)} & \textbf{69.6\(_{\textcolor{lightgray}{\pm2.7}}\)} & \textbf{77.2\(_{\textcolor{lightgray}{\pm1.2}}\)} & \textbf{62.8\(_{\textcolor{lightgray}{\pm0.8}}\)} & \textbf{76.7\(_{\textcolor{lightgray}{\pm0.5}}\)} & \textbf{63.9\(_{\textcolor{lightgray}{\pm0.5}}\)} & \textbf{71.8} \\ \bottomrule
\end{tabular}}
\end{table}

\subsection{Model Analyses}

\textbf{Robustness Analysis. }
We evaluate UAdapterGNN and full fine-tuning method under different noise levels to show the robustness of the proposed uncertainty-aware modeling. We randomly add or remove different proportions of edges to generate noisy graphs and then report the comparison results in Table~\ref{tab:robustness}. We can observe that: (1) UAdapterGNN consistently achieves better performance across all noise levels. (2) As the noise level increases, the relative improvement increases significantly. For example, in the random edge addition scenario, UAdapterGNN obtains a 4.90\% improvement at 80\% edge perturbation, substantially higher than the 0.66\% improvement at 20\% edge addition. These results demonstrate that incorporating uncertainty-aware modeling in adapter module can effectively enhance the robustness against graph noises.

\begin{figure*}[!htbp]
\centering 
\includegraphics[width=0.81\linewidth]{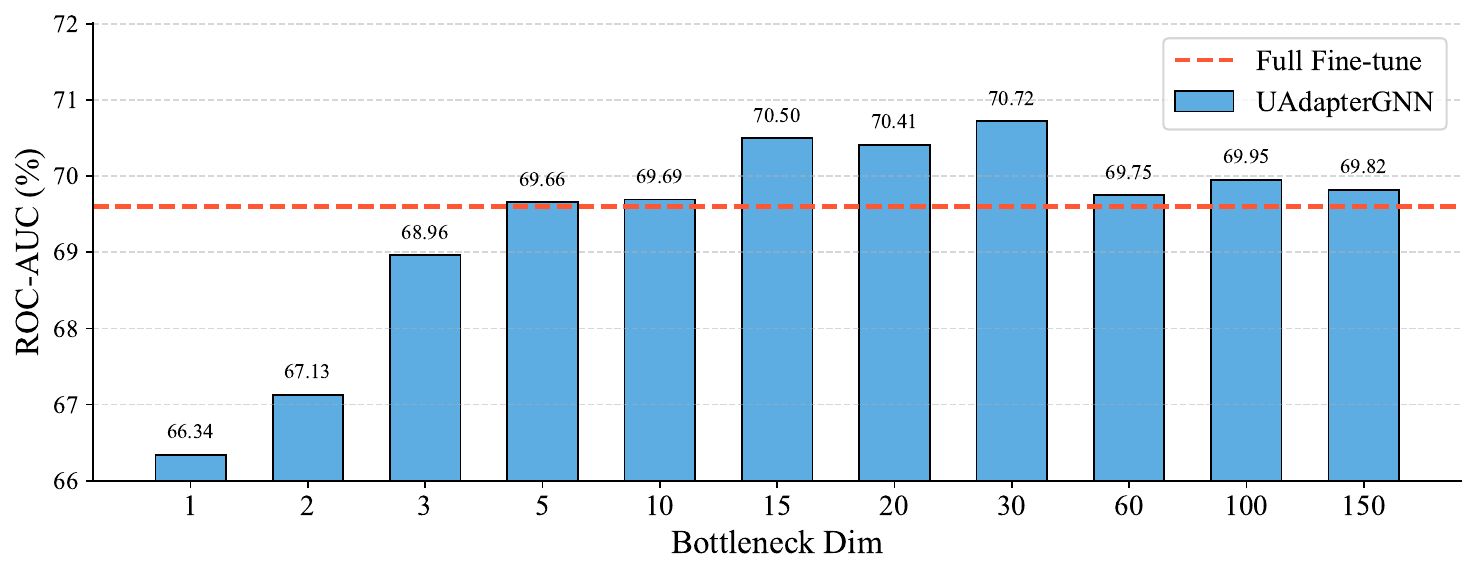}
\caption{Model performances with different bottleneck dimensions. Tuning the adapter’s bottleneck dimension restricts the tunable parameter space and enables knowledge transfer with very few parameters. If the bottleneck is too small, the model can under-fit and its performance will suffer.}
\label{performance_curve} 
\end{figure*}

\begin{figure*}[!htbp]
    \centering
    \includegraphics[width=\linewidth]{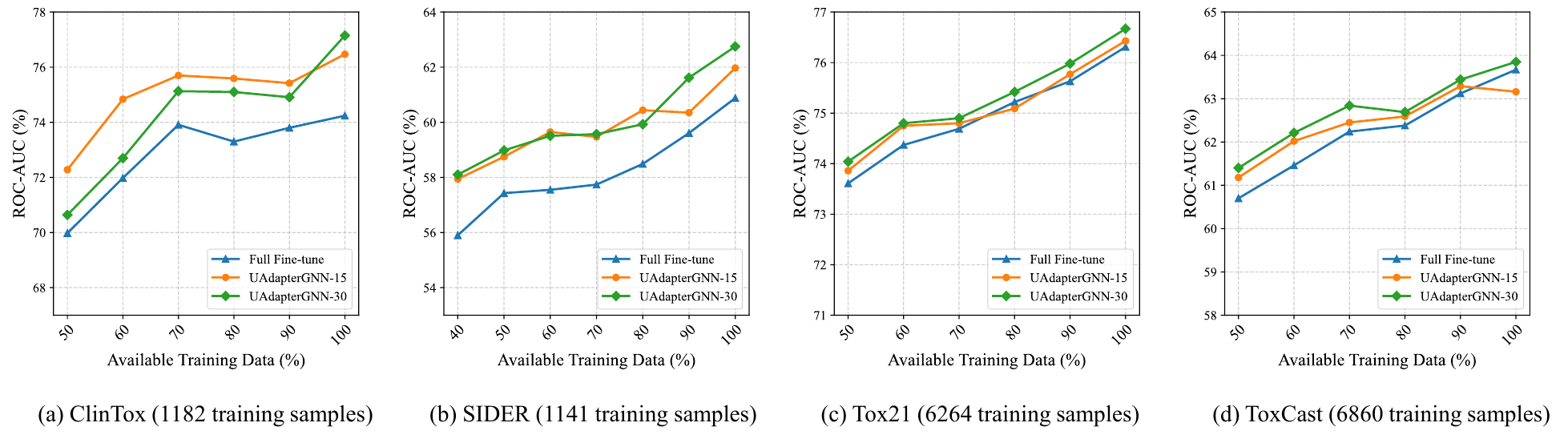}
    \caption{ROC-AUC (\%) performances with training data scaling up, compared to full fine-tuning. When data is scarce, UAdapterGNN with fewer tunable parameters  achieves better results.}
    \label{fig:train_ratio}
\end{figure*}

\textbf{Generalization Analysis. }
In this section, we investigate the generalization ability of UAdapterGNN on SIDER dataset. The gap between training and validation loss is commonly used to assess a model’s generalization ability~\cite{FengZDHLXYK020}, with a smaller gap indicating stronger generalization performance. We then derive a variant as a compared baseline by removing the Gaussion Random Adapter. As shown in Fig.~\ref{fig:sider_generalization}, the variant without the uncertainty module exhibits a notably larger gap, demonstrating the effectiveness of our uncertainty-aware adapter in enhancing the model’s generalization performance.

\textbf{Learnable Scaling Strategies. }
In our experiments, we adopt a learnable scaling strategy and compare it with fixed scaling factors $\{0.01, 0.1, 0.5, 1, 5\}$. As shown in Table~\ref{tab:scaling_ablation}, our method consistently outperforms all fixed settings across six datasets. We further observe that smaller scaling factors tend to better preserve pre-trained knowledge, whereas larger ones may lead to catastrophic forgetting. These results demonstrate that our learnable mechanism can dynamically achieve a balance between knowledge retention and parameter adaptation.

\textbf{Bottleneck Dimension Analysis. } It is noted that smaller bottleneck dimensions may restrict the model’s learning capacity while larger ones inevitably increase the parameter count.
To evaluate their effect in UAdapterGNN, we show the results across different dimensions, as shown in Fig.~\ref{performance_curve}. Note that the model achieves full fine-tuning performance with only 2.5\% of parameters at a dimension of $5$, while peak performance stabilizes within the range $[15, 30]$. These results suggest that the knowledge transfer from pre-trained model can be well achieved with minimal parameters by constraining the bottleneck dimension of the adapter.

\textbf{Training  Size Analysis. }
Here, we validate the performance of UAdapterGNN with different sizes of training data. Fig.~\ref{fig:train_ratio} demonstrates the comparison results of UAdapterGNN and full fine-tuned model. We can observe that our UAdapterGNN consistently outperforms the traditional full fine-tuning approach in all training settings. This indicates that the Gaussian random adapter in our UAdapterGNN can mitigate the overfitting issue in scenarios with limited training samples. 

\section{Conclusion}
This paper presents a novel fine-tuning approach, termed UAdapterGNN,  for Graph Neural Networks (GNNs) by 
 exploiting Gaussian random adapter to augment the
pre-trained GNN model. 
For the first time, the proposed UAdapterGNN effectively captures and utilizes uncertainty in graph data, leading to the improved  performance on fine-tuning various pre-trained GNN models. 
The integration of adaptive tuning allows the model to dynamically adjust based on data uncertainty, improving robustness and generalization  on the downstream tasks. Our experiments demonstrate that the proposed UAdapterGNN method consistently outperforms traditional GNN fine-tuning methods, validating its effectiveness in predictive accuracy, robustness and generalization ability. 


\bibliographystyle{IEEEtran}
\bibliography{reference}

\end{document}